\documentclass[letterpaper, 10 pt, conference]{ieeeconf}  
\pdfoutput=1

\IEEEoverridecommandlockouts                              

\usepackage{graphics} 
\usepackage{epsfig} 
\usepackage{mathptmx} 
\usepackage{times} 
\usepackage{amsmath} 
\usepackage{amssymb}  
\usepackage{lipsum,graphicx,subcaption,multicol}
\usepackage[firstinits]{biblatex}
\usepackage{hyperref}

\title{\LARGE \bf
Object Pose Estimation using Mid-level Visual Representations}

\author{Negar Nejatishahidin$^{*}$, Pooya Fayyazsanavi$^{*}$, Jana Ko\v{s}ecka
\thanks{These authors contributed equally to this work. George Mason University, e-mail:{nnejatis,pfayyazs,kosecka}@gmu.edu}
}

\begin{document}

\maketitle
\thispagestyle{empty}
\pagestyle{empty}

\begin{abstract}
This work proposes a novel pose estimation model for object categories that can be effectively transferred to previously unseen environments.
The deep convolutional network models (CNN) for pose estimation are typically trained and evaluated on datasets specifically curated for object detection, pose estimation, or 3D reconstruction, which requires large amounts of training data. In this work, we propose a model for pose estimation that can be trained with small amount of data and is built on the top of generic mid-level
representations~\cite{taskonomy2018}  (e.g. surface normal estimation and
re-shading). These representations are trained on a large dataset without requiring pose and object annotations. Later on, the predictions are refined with a small CNN neural network that exploits object masks and silhouette retrieval. The presented approach achieves superior performance on the Pix3D dataset~\cite{pix3d} and shows nearly 35\% improvement over the existing models when only 25\% of the training data is available.  We show that the approach is favorable when it comes to generalization and transfer to novel environments. Towards this end, we introduce a new pose estimation benchmark for commonly encountered furniture categories on challenging Active Vision Dataset~\cite{Ammirato2017ADF} and evaluated the models trained on the Pix3D dataset.

\end{abstract}

\section{Introduction}
Detecting objects and their 3D poses are an integral part of spatial 3D perception relevant to 
semantic simultaneous localization and mapping approaches~\cite{SLAM++} and target-driven navigation. 
The state of the art deep learning approaches have marked notable advancements by training
pose estimation models on large datasets with standard ResNet \cite{He2016DeepRL} backbone, requiring a large amount of training data and costly pose annotations. The resulting models did not generalize well to the the same instance in different environments. 

\begin{figure}[t]
\begin{center}
  \includegraphics[width=0.45\textwidth]{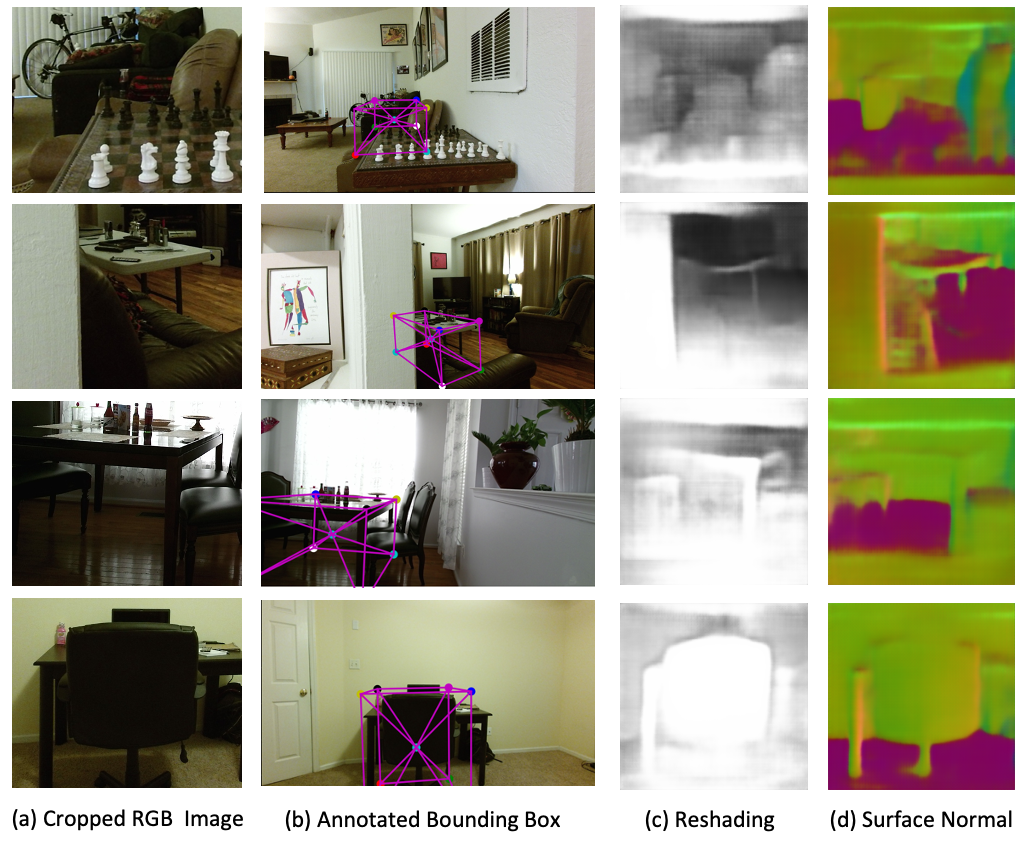}
\end{center}
   \caption{Active Vision Dataset~\cite{Ammirato2017ADF} pose estimation benchmark challenges; many objects are highly occluded (e.g. the first two rows), large lighting variations, glassy objects (the third row), and truncated viewpoints (last row). These challenges affect both computation of mid-level representations and pose estimation.}
    \label{AVD_sample}
\end{figure}

The proposed approach uses RGB images only and estimates the pose from a single view. The premise of the approach is to develop a method that can benefit from the availability of 3D CAD models, can be seamlessly integrated with the state-of-the-art object detection models, and will generalize well to novel environments. Instead of training the entire model end-to-end from pixels to pose predictions, we train a lightweight convolutional neural network on the top of generic mid-level representation features ({\em e.g.} normal estimation and re-shading features) that have been pre-trained in indoor environments. The initial predictions of the model are further refined using object masks and silhouette retrieval. The appeal of using mid-level representations features~\cite{midLevelReps2018} for this task is their re-usability and effectiveness for training visuomotor policies for exploration, navigation to target, as well as local planning and ability to transfer well to novel environments. 
In summary, the contributions of the proposed approach are summarized as follows:
\begin{itemize}
    \item Novel object pose estimation model build on the top of generic mid-level representation feature maps~\cite{taskonomy2018} (surface normals, and re-shading feature maps) that have been shown to be effective for other perceptual tasks. 
    \item Effective refinement stage aided by object detection masks and silhouette retrieval  achieving superior performance on the state-of-the-art Pix3D dataset~\cite{pix3d}.
    \item Competitive performance in low training data regime achieving 35\% improvement over the existing models when there's only 25\% of training data is available.
    \item New pose estimation benchmark for commonly encountered furniture categories on Active Vision Dataset~\cite{Ammirato2017ADF} and zero-shot pose estimation baseline for these real-world scans of indoor environments captured by the robot.
\end{itemize}


\section{ Related work}

There is a large body of work on 3D object pose estimation. 
The existing techniques vary depending on the sensing modality, focus on the object instances or categories, and availability of 3D models. With the success of deep convolutional neural networks (DCNN) for object recognition and detection, many works focused on estimating the pose (azimuth and elevation or full 6D pose) of object categories, separately or jointly with the object detection DCNNs by adding another branch to the network.  
In~\cite{mahendran20173d} a 3D pose regressor is learned for each object category, while in~\cite{mousavian20173d} a discrete-continuous formulation for the pose prediction is introduced, with the input being the cropped object bounding boxes. Authors in~\cite{xiang2017posecnn} decouple the pose estimation task into multiple components such as predicting pixel-wise object labels, estimating the object's center and distance from the camera to recover the translation, and estimating the rotation, while~\cite{poirson2016fast} and~\cite{kehl2017ssd} both extend the SSD~\cite{liu2016ssd} object detector to predict azimuth and elevation or the 6-DoF pose respectively. These methods learn the pose estimation and recognition directly from image pixels, and require a large number of training examples with pose annotations that are challenging to obtain for many categories. This is the case also for keypoint-based methods, which typically work better in the presence of occlusions~\cite{pavlakos20176, hueting2017seethrough}. 
Notable progress has been made in pose estimation for object instances from images, where 3D textured instance models were available during the training stage~\cite{Hodan2018BOPBF}.
Authors in~\cite{kundu20183d} exploit non-textured models where given the predicted pose and shape, the object is rendered and compared to 2D instance segmentation and trained end-to-end on the small number of categories. These approaches require 3D pose annotations in images during training~\cite{xiang2014beyond, hinterstoisser2012model}. Approaches that resort to keypoint based representations and use CAD models require annotations of 3D keypoints on textured CAD models. 
Authors in~\cite{li2017deep} generate a synthetic dataset provides additional supervision during training. They learn to predict the 2D image locations of the projected vertices  or projections of object's 3D bounding box ~\cite{tekin2018real} before a PnP algorithm estimates the pose.

At last, there are approaches that use point clouds or depth data to tackle the pose estimation problem. Examples of these include methods that exploit effective 3D shape representations of 3D objects~\cite{normalized-coord-CVPR19}, methods that learn how to align the sparse point clouds with the CAD models~\cite{Scan2CAD} or deep voting based methods~\cite{DeepSlidingShapes} that use point clouds both in training and testing. These works utilize repositories of 3D shape models~\cite{xiang2016objectnet3d,chang2015shapenet} and/or video or 3D datasets with pose annotations~\cite{KITTI, SUN-RGB-D}.

Our work is most closely related to the approach proposed in~\cite{pix3d}, where 2.5D sketches are learned as intermediate representations for 3D reconstruction and pose prediction. There the authors introduced a new benchmark for pose estimation with instance (instead of category) level annotations and adopted the 2.5D sketch representation~\cite{marrnet} as intermediate object representations for prediction of 3D structure and object pose. 
We take a departure from this training paradigm and show that effective pose estimation can be built on the top of feature maps that are part of generic perceptual skill set, also referred to as {\em mid-level representations}. The {\em mid-level representations}~\cite{midLevelReps2018} have been shown in previous works as useful priors for learning visuomotor policies. This work demonstrates their effectiveness for pose estimation task. 
\begin{figure*}[htbp]
\begin{center}
   \includegraphics[width=0.95\linewidth]{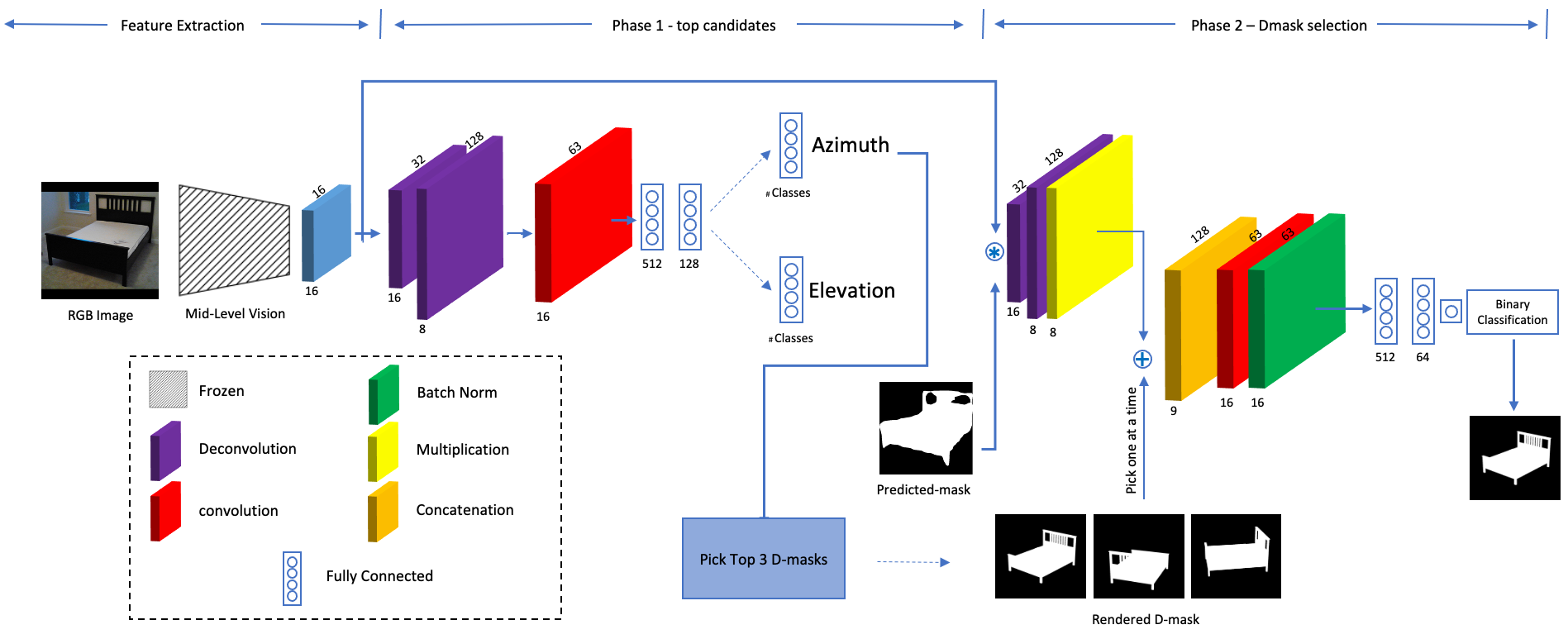}
\end{center}
   \caption{Network Architecture. First stage: up-sampled and fused mid-level feature maps (not the actual normal and re-shading images) are used to predict azimuth and elevation. Second stage: The object mask and mid-level features are compared with the top three object D-masks corresponding to most likely pose hypotheses from the first stage.}
    \label{Network}
\end{figure*}
\section{Approach}
In this work, we present a novel pose estimation approach using RGB images only. Our work is inspired by approaches that learn 2.5D sketches comprised of surface normal, depth map, and object silhouette as intermediate representations for 3D reconstruction and pose estimation~\cite{marrnet, pix3d}. Pix3D dataset comprised of several furniture categories, with instance-level pose and keypoint annotations along with depth, normal, and silhouette that are required for supervision of training the intermediate representations. 

The effectiveness and often the superior performance of the models trained in an end-to-end or on a single dataset usually comes with the generalization and domain adaptation challenges when applied in novel environments.
The fundamental question when it comes to building computer vision systems for robot perception is whether the existence of perceptual priors or representations learned through a set of proxy tasks ({\em e.g.} depth estimation, edge detection) can be useful for different downstream tasks.  Since a robot's visual perception requires tackling multiple tasks, the ability to share the representations/features that pertain to a particular class of environments is appealing. In this work, we propose to exploit generic mid-level representations~\cite{taskonomy2018} and show their effectiveness for down-stream object pose estimation task. 
The feature maps and models associated with mid-level representations are learned separately using a large number of images of indoor scenes with the proxy task supervision ({\em e.g.} re-shading and normal estimation) and are frozen in our approach.\newline
Our model consists of two stages. In the first stage, the initial pose predictions are made with up-sampled mid-level representations features to generate predictions for discretized candidate poses. For the top three pose candidates\footnote{$90\%$ of the time, the correct pose label is included in the top three pose candidates, which is optimized based on our experiments.}, we retrieve their discretized pose masks (Rendered D-Masks) and train a small neural network that takes the mid-level features representations gated (multiplied) by the object masks and learn to predict the correct pose of the retrieved mask. In this work, we will focus only on commonly encountered furniture categories and estimation of azimuth and elevation.
In the following, we will describe the two stages in more detail. 
\subsection{Mid-Level Visual Representations}
\label{input}
\begin{figure}[htbp]
\begin{center}

  \includegraphics[width=0.45\textwidth]{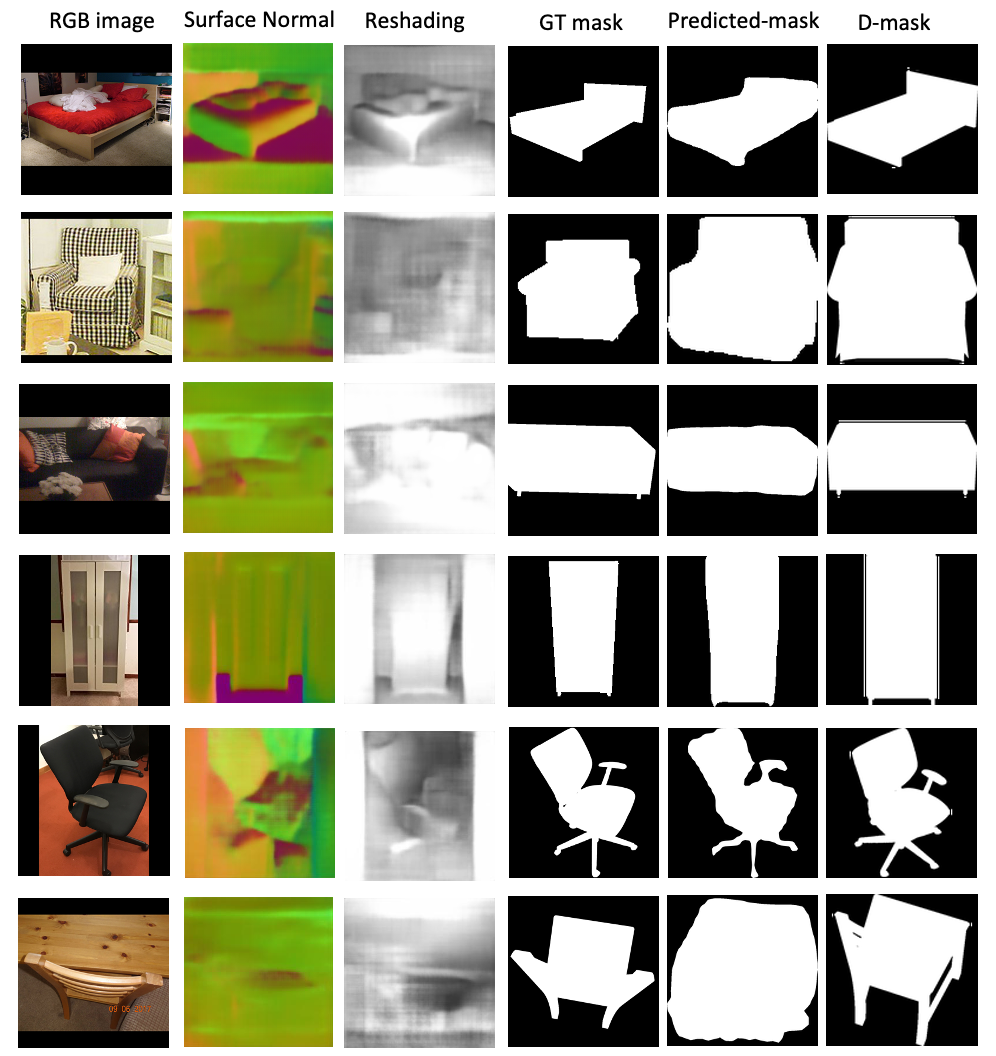}
\end{center}
   \caption{Visualization of mid-level representations and masks on the Pix3D dataset.}
    \label{inputmodel}
\end{figure}

The inputs to our model are the feature maps of mid-level representations trained separately
in indoor environments on visual proxy tasks in indoor environments~\cite{taskonomy2018}. Out of 25 available networks for different tasks, we found surface normal and re-shading features most effective for the pose classification model, see Figure~\ref{inputmodel}. We determined the most useful feature maps experimentally by testing their informative combinations and report results in the experiments section~\ref{mid_dif_input} in Table~\ref{mid-level-features}.  
These feature maps provide encoding of the input image, while the network weights are frozen, forming an input to our pose estimation model. We concatenate the $16 \times 16 \times 8$ feature maps from models trained on surface normal and re-shading tasks. Based on our experiments, using the actual predicted normal and re-shading images, same as pix3D \cite{pix3d}, reduces the performance of the model.
In the first phase of the model the feature maps are followed by additional convolution layers
to get initial azimuth and elevation predictions, see Figure~\ref{Network} (left). The predictions are further refined in the second stage using predicted object masks.





\begin{figure*}[htbp]
\begin{center}
   \includegraphics[width=1\linewidth]{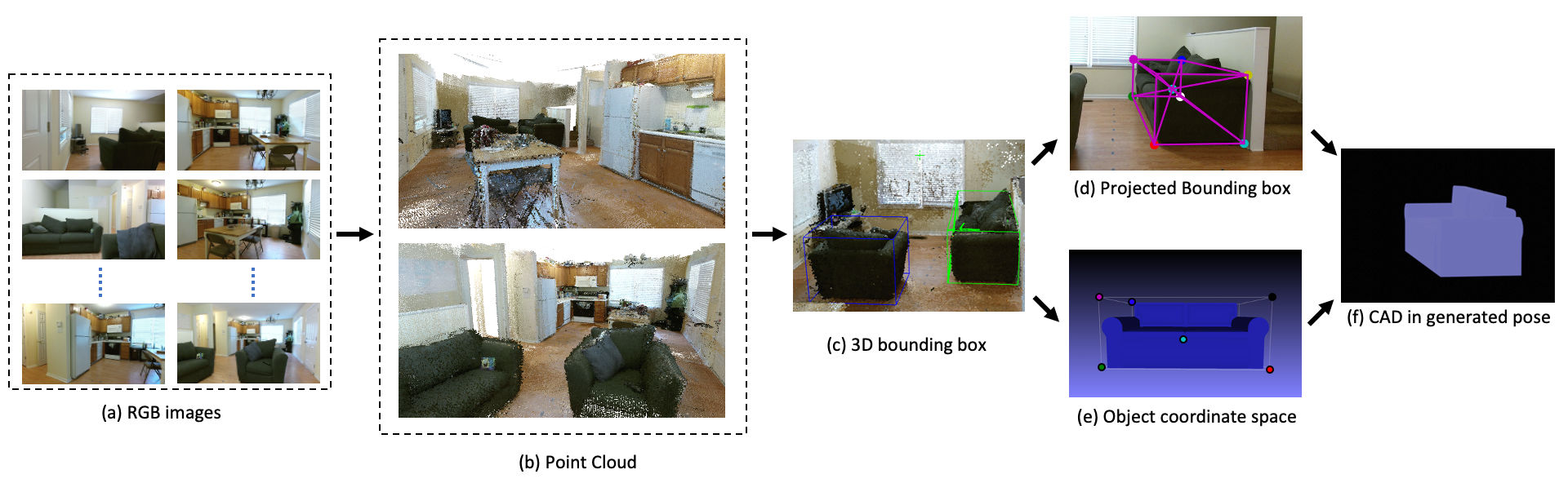}
\end{center}
   \caption{The pose labelling pipeline for main object categories in AVD. Using (a) RGB and depth images of each scene, we reconstructed the (b) dense 3D point-cloud of each scene. (c) The 3D bounding boxes of objects inside these point clouds are annotated using the LabelCloud tool. 
   (f) Poses are generated using the PNP algorithm between (d) the corners projected on the image plane and (e) corners in object coordinate frame. In total, we have labeled 6337 objects pose and 3D bounding box.}
    \label{AVD_labeling_pipline}
\end{figure*} 
\subsection{Masks and Discretized-Masks}
\label{dmask}
The second stage of our model aims to refine the predictions by using the predicted 
object masks, mid-level representations features, and rendered masks from CAD models.  
Since the state-of-the-art mask prediction models \cite{He2020MaskR,Nie2020Total3DUnderstandingJL} are not accurate enough to be solely effective, we generate masks rendered from different viewpoints of the CAD models of the training set and use them to further refine the predictions of the model from the first stage. We called these masks, Discretized-mask (D-mask). The 45 masks from different viewpoints, 9 different azimuths and 5 different elevations ($9 \times 5$) are stored per instance. In addition, for each image, we used the mask output of the state-of-the-art object detector~\cite{Gkioxari2019MeshR}, we call it predicted-mask, see Figure \ref{inputmodel}.
\subsection{Model}
\label{Model}
Figure \ref{Network} shows different stages of the approach. In the first stage, the model proposes its top candidates as the probable pose classes. The input to the model is only the surface normals and re-shading feature maps generated from mid-level visual representations, each map is $16 \times 16 \times 8$. We concatenate the features to get the input representation ($16 \times 16 \times 16$) and up-sample the concatenated features to the size of $128 \times 128 \times 8$. The up-sampled features are then passed to a convolutional layer. For the classification purpose, we use three fully connected layers, batch normalization, and ReLU, along with the dropout to get the final embedding before Softmax classification. Azimuth and Elevation are estimated separately from the last fully connected layer.
In the second stage, the up-sampled feature maps are masked out using the 'predicted-mask'. The top three pose candidates and their D-masks are stacked with the features one at a time.
These channels are followed by convolution layers along with batch normalization and three fully connected layers before the binary classification. The model in the second phase learns to determine whether the selected D-mask matches the features and the predicted-mask. Since our method uses only RGB images, in the testing stage, we need to retrieve correct CAD model to use it's D-masks. As a CAD model retrieval pipeline, we used the predicted-mask of the test image and compare with all the predicted-masks of the training data. For this stage, we found out that a simple Template Matching is sufficient to find the most similar CAD model for the pose estimation task, but more elaborate silhouette matching techniques, such as Chamfer matching can be used in practice. Although the CAD model retrieval part is time consuming, it is guided by the knowledge of object category provided by object detector and for many indoor setting applications, this needs to be done once for each object.




\section{Active Vision Dataset Pose Labeling}

The AVD dataset~\cite{Ammirato2017ADF} is a public dataset for active robotic vision tasks. It is comprised of dense scans of real indoor environments and has a total of 17 scenes. 
Each object is viewed from multiple viewpoints while the robot traverses the environment making it suitable for pose estimation. The pose estimation challenges include occluded objects, truncated images, dark objects, reflections, shiny objects, glass, lighting variations, and novel object instances. Figure \ref{AVD_sample} shows examples of these challenges. 
To train and evaluate pose estimation on AVD we
first provide pose annotation for the main object categories of {\em sofa, table, desk, bed}, and {\em chair}.
We first get the dense 3D dense point-cloud of each scene using each scene RGB and depth images and annotate the 3D bounding boxes for objects using 
LabelCloud tool~\cite{Sager2021labelCloudAL}, the example is in Figure \ref{AVD_labeling_pipline}c.
We then project corners of 3D bounding-boxes in world coordinate are projected back to the image plane using the transformation matrix from world to camera, Figure \ref{AVD_labeling_pipline}d, base on the following equation: $$ 
{\bf X_w}  = [R_c^w|T_c^w] {\bf X_c}
 $$
\begin{figure}[htbp]
\begin{center}
   \includegraphics[width=\linewidth]{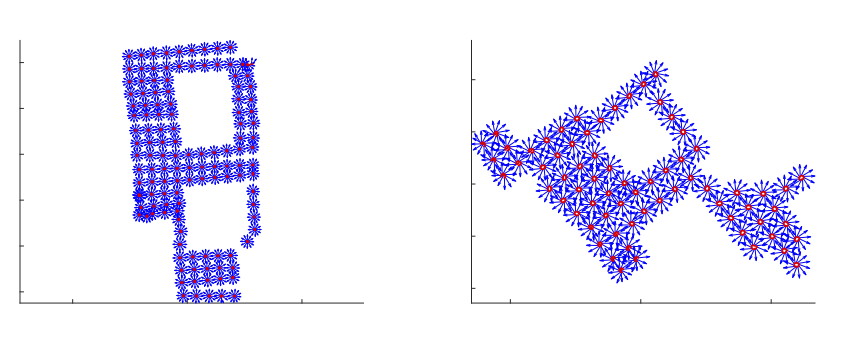}
\end{center}
   \caption{The image shows the dense RGB-D sampling from each home in the AVD dataset. The red dots are the locations of the camera, and the blue arrows around each dot are the 12 camera orientations.}
    \label{vew_AVD}
\end{figure}
considering that, $R_c^w = {R_w^c}^T$, and $T_c^w = {-R_w^c}^{T} T_w^c$, ${\bf X_c} = [X_{c}, Y_{c}, Z_{c}]^T$ is a point in the camera coordinate frame and ${\bf X_w} = [X_w, Y_w, Z_w] ^T$ is the point in the world coordinate frame. $R_w^c$ is the rotation matrix from the world to the camera and $T_w^c$ is the translation between them. The camera intrinsic parameters matrix $K$ is also known. See Figure \ref{AVD_labeling_pipline}b for an example of dense point-cloud reconstruction. 
To get the pose of an object in camera coordinate frame we used PnP algorithm~\cite{Lepetit2008EPnPAA} between the points in 2D image frame and correspondences in the object coordinate frame. The 3D bounding box can be defined by it's center $c = [c_x,c_y,c_x]$ which in object coordinate is $[0,0,0]$, it's orientation $R(\theta, \phi, \alpha )$, and it's dimensions $D = [d_x,d_y,d_z]$ which is annotated in second step. Therefore, the corners in object coordinate are ${\bf X}_1^b = [ d_x/2, d_y/2,d_z/2]^T$, ${\bf X}_2^b = [ - d_x/2, d_y/2,d_z/2]^T$, $...$, ${\bf X}_8^b = [ - d_x/2, -d_y/2,-d_z/2]^T$. The example object coordinate space is shown in Figure \ref{AVD_labeling_pipline}e.
In the last step, the PnP algorithm~\cite{Lepetit2008EPnPAA} is used to estimate rotation and translation of the object's 3D bounding-box from correspondences between the 2D image points and 3D points in the object coordinate frame. The result is shown in Figure \ref{AVD_labeling_pipline}.

\begin{table*}
\centering
\vspace{6pt}
\scalebox{0.95}{
\begin{tabular}{|c|c|c|c|c|c|c|c|c|c|c|c|}
\hline
 &\multicolumn{1}{c|}{bed} &\multicolumn{1}{c|}{chair} & \multicolumn{1}{c|}{desk} &\multicolumn{1}{c|}{table} &\multicolumn{1}{c|}{sofa} & \multicolumn{1}{c|}{wardrobe} &\multicolumn{1}{c|}{bookcase} &\multicolumn{1}{c|}{misc} & \multicolumn{1}{c|}{tool} &\multicolumn{1}{c|}{mean Azimuth} & \multicolumn{1}{c|}{mean Elevation} \\
\cline{1-10}
\hline\hline
Baseline & 65.73 & 50.73 & 41.56 & 51.08 &  78.55 &  77.78 & 65.82 & 20.00 & 18.18 & 56.72 &  69.45\\ \hline
ResNet\_baseline  & 68.08 & 58.95 & 50.65 &51.80 &  81.20 & 72.22 & 67.09 & 20.00 & 27.27 & 61.78 &  75.25\\ \hline
Mousavian et al. & 71.83 & 47.99 & 51.30 & 56.59 & 80.96 & 75.93 & 68.35 & 15.00 & 18.18 & 57.87 &  75.81\\ \hline
Ours\_phase1 & 75.12& 70.61 & \textbf{62.34} &60.43&  \textbf{88.19}&  \textbf{94.44} & \textbf{84.81} & \textbf{40.00} & 27.27 & 72.21 &  \textbf{76.08} \\ \hline
Ours &  \textbf{77.93}&  \textbf{74.55} &  59.09 &  \textbf{61.39} &  85.06&  90.74 & 83.54 & 30.00 &\textbf{36.36}& \textbf{73.56} & \textbf{76.08} \\ \hline
\end{tabular}
}
\caption{The azimuth classification accuracy for $9$ bins with $2.5^\circ$ overlap. Models are all trained category agnostic. }
\label{category_level}
\end{table*}
\section{Experiments}
We train and evaluate the proposed pose estimation approach on the Pix3D dataset~\cite{pix3d} comparing it to the state-of-the-art approaches \cite{pix3d,mousavian20173d} and some baseline methods. We then directly evaluate it without any additional training or fine-tuning on the challenging AVD dataset\cite{Ammirato2017ADF} to see the generalizability of the approach. Code is also available at \url{https://github.com/N-NEJATISHAHIDIN/Pose_from_Mid-level}.



\subsection{Training} 
We use frozen mid-level-representation features as the input to our model,
and set the learning rate to $0.001$ with a step size of $3$. All the models are trained for $10$ epochs with early stopping. The first phase is trained with {\em cross-entropy} loss, where azimuth $\theta$ and elevation $\phi$ are the logits over the number of bins with 
$L_{\theta, \phi}=L_{\theta}+L_{\phi}$
The second phase is trained with binary cross entropy loss (BCE) between the target and the output logits.
$$
L_{mask}=-\frac {1}{n} \sum_{i=0}^{n} y_{i}^* \cdot \log \left (y_{i}\right )+\left (1-y_{i}^*\right ) \cdot \log \left(1-{y}_{i}\right )
$$
where $y_i^* = 1$ when the chosen
D-mask is from the ground truth bin.
Since Pix3D does not provide a train/test split, we use widely adopted Mesh R-CNN split S1. Split S1 randomly allocates $7539$ images for training and $2530$ for testing. The model is trained in the category agnostic level, 
{\em i.e.} we have only a single model for all categories.

\subsection{Testing} 
\begin{figure}[htbp]

\begin{center}
   \includegraphics[width=0.95\linewidth]{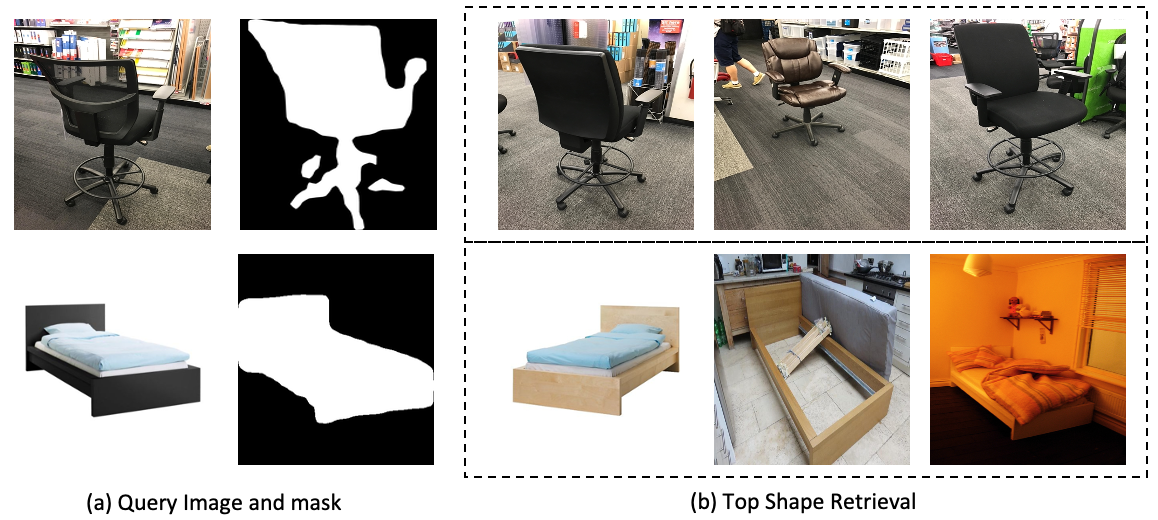}
\end{center}
   \caption{Shape Retrieval on Pix3D. Template matching compares the predicted-mask of the query image with other predicted-masks in the training data of the same category.}
    \label{shaperetv}
\end{figure}

For every test image, we find the nearest CAD model using template matching from OpenCV. To find the nearest CAD model, we use the predicted-mask of the image and compare it with the training predicted-masks of the images from the same category. To be more robust and scale-invariant, we compare each mask at different scales. See sample results are in Figure~\ref{shaperetv}.
As discussed in Section~\ref{Model}, we use the top three pose candidate bins from stage one along with their probability. 
At testing time, we run binary classification between each of the top three candidate poses D-masks and the predicted\_mask from the image. If the predicted\_mask and D\_mask are in the same bin, stage two output is a positive match. In case of more than one positive output, we choose the bin which maximizes the $\max_{\forall i \in C_{pose}}{\hat{Y}_i P_{phase1}({i})}$, where $C_{pose}$ is the set of three candidate poses from phase one, the $\hat{Y}_i$ is the binary classification result for bin $i$ from phase2, and $P_{phase1}({i})$ is the output probability of the bin from phase1. 



\subsection{Pose Evaluation}
\subsubsection{Baseline vs. Ours}
%
To establish the importance of mid-level representations, we performed an ablation experiment in which the same network architecture is used but trained entirely from scratch. We called this network Baseline. Our second baseline, ResNet\_baseline, uses ResNet backbone features instead of the features of the mid-level representation. The results for phase one and Our whole model are shown in Table~\ref{category_level}. The backbone from mid-level features in our model is frozen. 


\begin{table}
\centering
\scalebox{0.95}{
\begin{tabular}{|c|c|c|}
\hline
 &\multicolumn{1}{c|}{Azimuth} &\multicolumn{1}{c|}{Elevation} \\

\hline\hline
Normal & 68.37 & 74.47  \\
\hline
Normal+ 2D key-points & 68.77 & 74.38    \\
\hline
Normal+edge\_occlusion  & 69.52 & 75.91  \\
\hline
Normal+ depth\_euclidean  & 70.24 & \textbf{76.36}  \\
\hline
Normal+ segment\_unsup25d  & 70.40 & 74.42  \\
\hline
Normal+ 3D key points & 71.03 & \textbf{76.36} \\
\hline
Normal+reshading & \textbf{72.21} & 76.08 \\
\hline
\end{tabular}
}
\caption{The choice of mid-level features and the effect on pose classification task with $9$ bins with $2.5^\circ$ overlap.}
\label{mid-level-features}
\end{table}

\subsubsection{Mid-level representations features input}
\label{mid_dif_input}
The $16 \time 16 \times 8$ feature maps of mid-level representations are used as our model input. 
Inspired by~\cite{marrnet, pix3d} we use the surface normals that have been shown to be beneficial for the pose estimation. From other $24$ models introduced in~\cite{taskonomy2018}, the re-shading feature map benefits our model most as shown in our experiments in Table~\ref{mid-level-features}. Figure~\ref{inputmodel} shows sample inputs. We also tried the combinations of $3$ feature maps, but they did not improve the model by more than a few $0.1\%$. 

The second stage of the approach completely relies on the D-Masks. With the even number of bins, the symmetric objects have the same D-masks for multiple orientations. This fact encourages us to use the odd number of bins. Based on our experiments, it's difficult for models to distinguish the pose of the object if it's close to the bin's borders, hence overlapping bins are used as in \cite{mousavian20173d}. 
To make the results comparable to other pose classification models, we defined $9$ bins with $2.5$ degree overlap on each side; which is comparable to $8$ bins. Since the range of each bin in both is $45$ degree; Similarly, the $5,13,25$ bins with $9,1.15,0.3$, degree overlap on each side is comparable with $4,12,24$ bins. Table~\ref{difrent_bins} shows the model performance for the different number of bins.

\subsubsection{Comparison with Available Models}
In comparison Mousavian et al. \cite{mousavian20173d}, our model outperforms by a large margin, which shows the importance of generalizability of the mid-level features for the pose estimation task, Table \ref{category_level}. The model is also compared with the Pix3D dataset~\cite{pix3d} on the chair category. The Pix3D trained reconstruction and pose estimation together only for the untruncated and unoccluded chairs. It is stated that the reconstruction branch helps to improve the pose classification accuracy. Compared to Pix3D, our model is trained category agnostic way, with no need for supervision on the mid-level features. On the contrary, Pix3D needs supervision for normal, depth, and silhouette images and is trained per category. The results are shown in Table~\ref{compare_with_pix3d} for the chair category.




\subsubsection{Low Data Regime for training}
\begin{figure}[htbp]
\begin{center}
   \includegraphics[width=\linewidth]{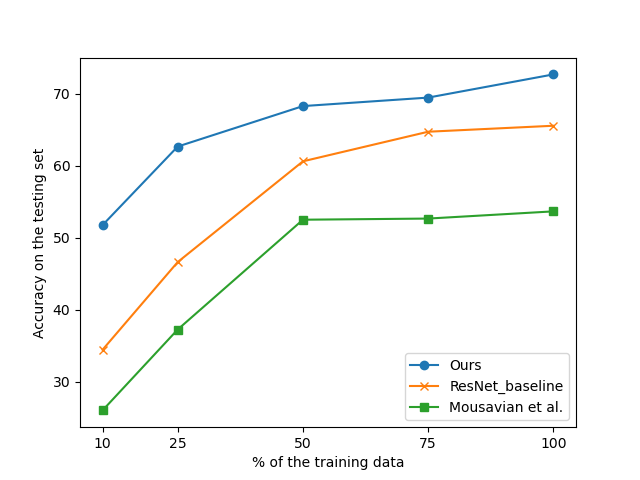}
\end{center}
   \caption{The accuracy results with training on different percentages of the training data. Achieving 35\% improvement over the existing models when there's only 25\% of training data is available.}
    \label{less-data}
\end{figure}

Training Models for pose estimation tasks from scratch require thousands of samples for each category with the pose annotation. Some approaches \cite{pix3d} also require the surface normal, silhouette, or depth supervision. The labeling process is costly and time-consuming. The proposed model outperforms other models by a large margin when using a fraction of training examples. Figure \ref{less-data} shows some quantitative results.

\begin{table}
\centering
\scalebox{0.88}{
\begin{tabular}{|c|c|c|c|}
\hline

\# bins & $(4) / (5 + 9^\circ)$ & $(8) / ( 9+ 2.5^\circ)$  & $(12 ) / (13+ 1.5^\circ)$ \\
\hline
\hline
Pix3D & $76.00$ & $73.00$ & $61.00$    \\
\hline
ours & \textbf{$76.61$} & \textbf{$74.55$} & \textbf{$61.70$}    \\
\hline
\end{tabular}
}
\caption{The comparison of our model with the Pix3D model. Results are only for the chair category for different number of bins.}
\label{compare_with_pix3d}
\end{table}

\begin{table}
\centering
\scalebox{0.88}{
\begin{tabular}{|c|c|c|c|c|c|c|}
\hline
 &\multicolumn{3}{c|}{Azimuth} &\multicolumn{3}{c|}{Elevation} \\
\hline
\# bins & $5 + 9^\circ$ & $9+ 2.5^\circ$  & $13+ 1.5^\circ$  & $3$ & $5$ & $7$  \\
\hline
\hline
ours & $77.54$ & $73.56$ & $64.82$ & $85.37$ & $76.08$ & $66.36$    \\
\hline
\end{tabular}
}
\caption{The average accuracy of the azimuth and elevation for the different number of bins.}
\label{difrent_bins}
\end{table}

\subsection{AVD experiments}
\begin{figure}[htbp]

\begin{center}
  \includegraphics[width=\linewidth]{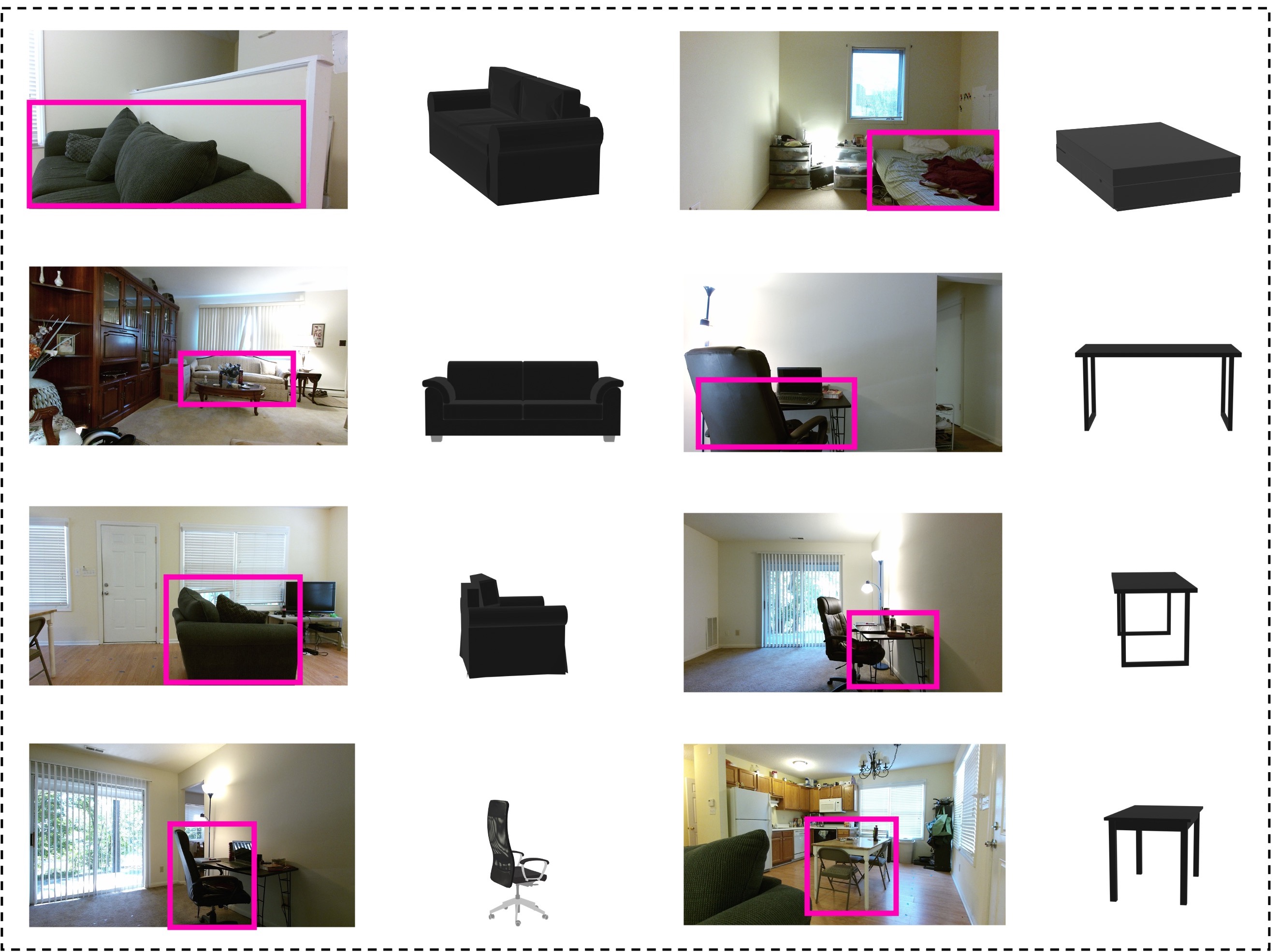}
\end{center}
  \caption{Results on the challenging AVD dataset. This model is capable of estimating pose without any training on AVD data.}
    \label{ssresults}
\end{figure}

As mentioned earlier, the AVD dataset is a challenging dataset for pose estimation task; most of the images are either truncated or heavily occluded. The generated normal and re-shading features for these images are less accurate in comparison to the Pix3D, see Figure~\ref{AVD_sample}. The fact that AVD has been densely sampled from different views, generates object poses that are less probable in the Pix3D, such as the back of the sofa. Figure~\ref{vew_AVD} is the map of two homes with the location and orientation of the camera. Since the predicted masks for AVD are of lower quality because of the mentioned challenges, we report the results for the Our\_phase1 model. The model was trained on Pix3D and tested on AVD. We also evaluated the ResNet\_baseline model performance on AVD to better show the generalizability of our model. Table~\ref{AVD_category_level_results} shows the results. The proposed Ours\_phase1 outperforms the other model with $10.56\%$. This shows the effectiveness and generalizability of mid-level features. In comparison to Pix3D, our model's failures include objects in rare poses or cases of heavily occluded objects. Figure \ref{ssresults} shows some of the results on the AVD dataset.

\begin{table}
\vspace{8pt}
\centering
\scalebox{0.95}{
\begin{tabular}{|c|c|c|c|c|c|c|}
\hline
Category &\multicolumn{1}{c|}{bed} &\multicolumn{1}{c|}{chair} & \multicolumn{1}{c|}{desk} &\multicolumn{1}{c|}{table} &\multicolumn{1}{c|}{sofa} 
&\multicolumn{1}{c|}{mean}\\
\hline
Total \# samples & 338 & 2026 & 313 & 1435 & 2225 & 6337 \\
\hline\hline
ResNet\_baseline (\%) &  13.31 & 41.06 & 25.88 & 22.43 & 44.04 & 35.66\\
\hline
Ours\_phase1 (\%) & 23.08 & 46.54 & 45.04 &  29.97 &60.43 & 46.22 \\
\hline
\end{tabular}
}
\caption{Per category accuracy of two models on AVD dataset. The results are for 9 bins with $2.5^\circ$ overlap. }
\label{AVD_category_level_results}
\end{table}


\section{Conclusions}
We present a novel object pose estimation approach built on the top of generic mid-level representations trained on computer vision proxy tasks
of surface normal estimation and reshading. The first stage is competitive with the state-of-the-art approaches that are trained on the Pix3D dataset. The second refinement via learned retrieval stage achieves superior performance compared to the state-of-the-art. 
We also introduce a new pose estimation benchmark on the Active Vision Dataset and establish several new pose estimation baselines. We currently formulate the problem as classification and estimate only discretized azimuth and elevation angles. We plan to address the full 6 DOF pose estimation in the future. The performance of our approach is notably affected by the quality of the object masks, which deteriorates with challenging viewpoints and occlusions.


{\renewcommand*{\bibfont}{\footnotesize}
\printbibliography
}

\end{document}